\documentclass[letterpaper]{article} % DO NOT CHANGE THIS
\usepackage{aaai2026}  % DO NOT CHANGE THIS
\usepackage{times}  % DO NOT CHANGE THIS
\usepackage{helvet}  % DO NOT CHANGE THIS
\usepackage{courier}  % DO NOT CHANGE THIS
\usepackage[hyphens]{url}  % DO NOT CHANGE THIS
\usepackage{graphicx} % DO NOT CHANGE THIS
\usepackage{multirow}
\usepackage{natbib}  % DO NOT CHANGE THIS AND DO NOT ADD ANY OPTIONS TO IT
\usepackage{caption} % DO NOT CHANGE THIS AND DO NOT ADD ANY OPTIONS TO IT
\usepackage{algorithm}
\usepackage{algorithmic}

\usepackage{newfloat}
\usepackage{listings}
\DeclareCaptionStyle{ruled}{labelfont=normalfont,labelsep=colon,strut=off} % DO NOT CHANGE THIS
\floatstyle{ruled}
\newfloat{listing}{tb}{lst}{}
\floatname{listing}{Listing}
\title{EraserDiT: Fast Video Inpainting with Diffusion Transformer Model}
\author{
    Jie Liu,
    Zheng Hui
}
\affiliations{
    \textsuperscript{\rm 1}Mango TV\\
    Changsha, Hunan 410083 China\\
    jieliu543@gmail.com,
    zheng\_hui@aliyun.com
}

\usepackage{bibentry}
\begin{document}

\maketitle

\begin{abstract}
Video object removal and inpainting are critical tasks in the fields of computer vision and multimedia processing, aimed at restoring missing or corrupted regions in video sequences. Traditional methods predominantly rely on flow-based propagation and spatio-temporal Transformers, but these approaches face limitations in effectively leveraging long-term temporal features and ensuring temporal consistency in the completion results, particularly when dealing with large masks. Consequently, performance on extensive masked areas remains suboptimal. To address these challenges, this paper introduces a novel video inpainting approach leveraging the Diffusion Transformer (DiT). DiT synergistically combines the advantages of diffusion models and transformer architectures to maintain long-term temporal consistency while ensuring high-quality inpainting results. We propose a Circular Position-Shift strategy to further enhance long-term temporal consistency during the inference stage. Additionally, the proposed method interactively removes specified objects, and generates corresponding prompts. In terms of processing speed, it takes only 65 seconds (testing on one NVIDIA H800 GPU) to complete a video with a resolution of 2160 × 2100 with 97 frames without any acceleration method. Experimental results indicate that the proposed method demonstrates superior performance in content fidelity, texture restoration, and temporal consistency. Project page: \small{\url{https://jieliu95.github.io/EraserDiT_demo/}}.
\end{abstract}


\section{Introduction}
Video inpainting (VI) aims to synthesize appropriate content that is visually realistic, semantically accurate, and temporally consistent. This technique allows content creators to remove unwanted objects from original videos. Previously, mainstream VI methods have included flow-guided propagation and video Transformer.

The flow-guided propagation series~\cite{zhou2023propainter} typically includes flow completion, feature propagation, and content generation. The precision of flow completion directly impacts the quality of the final result. \emph{ProPainter}~\cite{zhou2023propainter} adopts a recurrent flow completion to provide precise optical flow fields for subsequent propagation modules. Equipped with dual-domain propagation and mask-guided sparse Transformer, \emph{ProPainter} effectively performs propagation at the feature level between adjacent frames. However, when encountering large areas of masking, the generative capability of Generative Adversarial Networks (GANs) has been proven to be insufficient.

\begin{figure}
	\centering
        \includegraphics[width=0.50\textwidth]{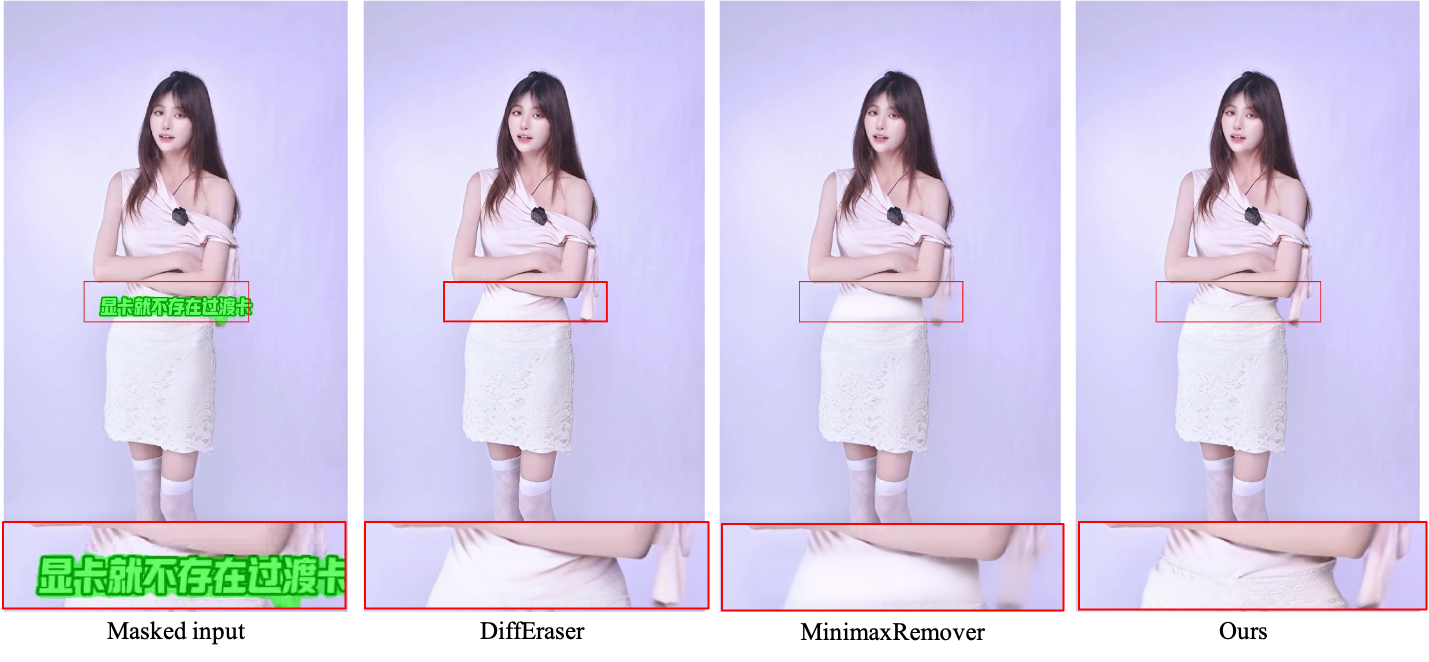}
	\caption{The visual results of DiffEraser, MinimaxRemover, and the proposed method. DiffEraser lacks texture and fine details. MinimaxRemover removes the target but leaves visible edge artifacts. Our method preserves more texture and detail, especially at the junction between the top and the skirt.}
	\label{fig:demo_text}
\end{figure}

Recently, diffusion models have gained popularity among researchers due to their superior performance in image and video generation. Li et al.~\cite{li2025diffueraser} explored a video completion scheme based on the diffusion model \cite{guo2023animatediff,zi2025minimax}, attempting to address situations that previous methods struggled with (such as completing large gaps). They incorporate prior knowledge (inpainted by \emph{ProPainter}) into the diffusion model to perform noise initialization. However, limited by the video generation capabilities of Minimax-Remover, its ability to complete videos with natural texture is restricted. As shown in Figure~\ref{fig:demo_text}, this usually manifests itself as regular texture artifacts and blurries. DiffEraser lacks texture and semantic coherence. The output looks unnatural. MinimaxRemover removes the target but breaks structural continuity. Edge artifacts are visible. Our method ensures visual naturalness. Semantic consistency is preserved. Clothing transitions appear smooth and realistic.

Recently, text-to-video tasks based on diffusion transformer (DiT) models~\cite{yang2024cogvideox,hacohen2024ltx,kong2024hunyuanvideo} have demonstrated highly commendable generation results. Unlike previous methods that use separated spatial and temporal attention to reduce computational complexity, these methods employ a 3D full attention mechanism to maintain the consistency of large-movement objects. These methods adopt 3D Variational Autoencoders (3D VAE) to implement video compression. Compared to the 2D VAE used by Animatediff, it not only compresses along the temporal dimension but also reduces the flickering issues in generated videos. 

\begin{figure*}
	\centering
        \includegraphics[width=0.95\textwidth]{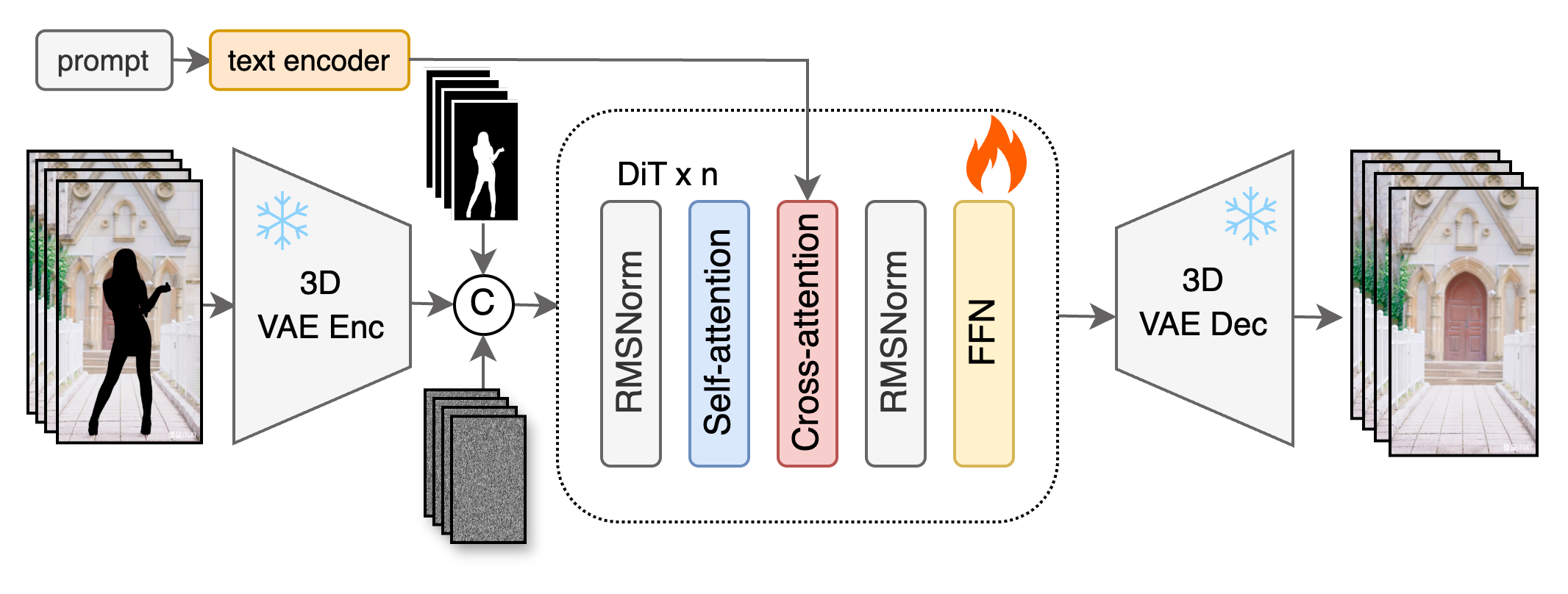}
	\caption{The training pipeline of the proposed method for video object removal. In this method, the pipeline requires three inputs: a masked video sequence, a mask video, and a corresponding brief text prompt. The masked video sequence undergoes encoding by the VAE to obtain latent features. These features are then concatenated with the noise and the downsampled mask sequence in the spatio-temporal domain along the channel dimension. During the training phase, fine-tune all the parameters of the video transformer while keeping the other parameters frozen.}
	\label{fig:train_pipeline}
\end{figure*}

Consequently, leveraging the powerful generative capabilities and spatial-temporal consistency of video DiT, exploring large-gap and large-motion video inpainting is a feasible direction.

In this study, we developed an interactive video completion method based on video DiT architecture to handle large object removal tasks in high-resolution (up to 1080p) and large-motion scenes. Leveraging the powerful generative capabilities of the foundational text-to-video large model, we can remove various types of objects, including people, animals, plants, vehicles, subtitles, and other common targets. The main contributions of this work can be summarized as follows. 

% itemize
\begin{itemize}

\item This paper introduces DiT for maintaining spatiotemporal consistency in high-resolution videos during object removal tasks. DiT employs 3D full attention to enhance spatiotemporal consistency and leverages the high downsampling scale of the 3D Causal VAE to efficiently process high-resolution video object removal up to 1080p.
\item To better maintain temporal consistency, we propose an inference method called the Circular Position-Shift strategy.
\item To facilitate the removal of large objects in high-resolution videos (up to 1080p) with significant motion, this work generated approximately 60,000 mask videos for dynamic objects, including people and animals. 
\item To improve usability and achieve more accurate video descriptions during the testing phase, we propose an automated method for generating text prompts for videos that do not contain the object to be removed. This approach guarantees a more efficient and precise alignment between the video content and the descriptive text.
\end{itemize}

\section{Related Work}
\label{sec:related}

%-------------------------------------------------------------------------

\subsection{Video Inpainting}
The current mainstream video inpainting methods can be broadly categorized into Flow-Guided Propagation-based and Video Transformer-based approaches. Flow-guided propagation has become a pivotal technique in video inpainting, leveraging completed optical flow to align frames seamlessly while ensuring temporal coherence throughout the video.

Turning attention to Video Transformers, a groundbreaking approach in this field is STTN~\cite{zeng2020sttn}, which incorporates ViT~\cite{vit} into video inpainting, providing an effective solution to address both spatial and temporal dimensions simultaneously. Liu~et al.~\cite{liu2021fuseformer} advanced this approach by refining transformer-based methods, improving the sub-token fusion ability of transformers for learning fine-grained features to better relicate the dynamics in video sequences. 

Among these methods, Propainter~\cite{zhou2023propainter} emerges as a notable approach, featuring recurrent flow completion, dual-domain propagation, and a mask-guided sparse Transformer. This method excels at propagating known pixels across all frames and showcases an early capability to synthesize unknown pixels. However, its generative capacity faces limitations when handling large masks, often resulting in visible artifacts. And RGVI \cite{cho2025elevating} integrates a large generative model with an efficient pixel propagation technique to achieve high-quality, controllable, and high-resolution video inpainting. MiniMax-Remover \cite{zi2025minimax} is a two-stage video object removal method that simplifies the model architecture and employs minimax optimization to achieve fast, high-quality results without relying on text input or classifier-free guidance. However, the inference time for a single 2K-resolution clip exceeds 7 minutes.

% \subsection{Video Editing}
% Video editing with text prompt

\subsection{Video Generation with DiT}
By leveraging Transformers as the foundation for diffusion models, specifically Diffusion Transformers (DiT), text-to-video generation has achieved a remarkable new milestone. Early notable work in the open-source community includes CogVideoX~\cite{yang2024cogvideox}, which introduces 3D full attention to replace the separated spatial and temporal attention, effectively resolving inconsistency issues
in the generated videos. CogVideoX primarily utilizes the multi-modal DiT (MM-DiT) architecture~\cite{sd3} and incorporates the Expert Adaptive Layernorm to independently handle each modality. HunyuanVideo~\cite{kong2024hunyuanvideo} integrates a combination of multi-stream DiT and single-stream DiT architectures, successfully training a video generation model with over 13 billion parameters. This model surpasses the performance of previous state-of-the-art models, such as Runway Gen 3 and Luma 1.6. HunyuanVideo's exceptional generative performance is commendable. However, it requires significantly more GPU memory consumption (60GB needed for 129 frames at 720p resolution) and has slower inference speeds. Recently, the Lightricks Team proposed LTX-Video~\cite{hacohen2024ltx}, a real-time video latent diffusion method. Its greatest advantage is its inference speed, thanks to the design of a $1:192$ compression ratio with spatiotemporal downsampling of $32\times32\times8$, enabling the generation of high-resolution video at unprecedented speed. Although LTX-Video's performance is not as good as HunyuanVideo, it offers a better balance between performance and speed, and the strong reference nature of video completion tasks means that the generative capability requirement is lower compared to pure text-to-video tasks. Therefore, we chose LTX-Video as our backbone model.

\section{Methodology}

To overcome these limitations, we propose a novel algorithm based on the Diffusion Transformer (DiT), which leverages simple prompts to significantly enhance video object removal. This approach adeptly addresses three key challenges: enabling the direct removal and completion of objects in 1080p resolution videos to maintain high-quality and detailed results; given the box of the object to remove, a prompt without that object can be automatically created; and effectively completing large masked regions, ensuring the superior removal of humans and other common targets. By combining the strengths of diffusion models and transformer architectures, our method not only improves long-term temporal consistency and quality inpainting but also establishes a robust and efficient solution for complex video inpainting challenges, advancing the state-of-the-art in video object removal and completion tasks. The specific methods are comprehensively detailed in the following subsections. 

\subsection{Training pipeline}
In the task of video object removal, both speed and quality are crucial. To address this, we have chosen \emph{LTX-Video} \cite{hacohen2024ltx} as our foundational architecture. As illustrated in Figure \ref{fig:train_pipeline}, this architecture integrates a VAE, a text encoder, and a denoising video transformer, whose inputs are a masked video sequence, the corresponding mask sequence, and the corresponding text prompt. During the training stage, we randomly select a video $V_{i}$ from various videos $V=\left \{ V_{1}, V_{2}, V_{3}, \dots ,V_{n}  \right \} $ and a mask video $M_{j}$ from various masks $M=\left \{ M_{1}, M_{2}, M_{3}, \dots ,V_{n}  \right \} $. These videos are then fed into the model for training.

The latent features are extracted from the masked video sequence $V_{M} = V_{i} * M_{j}$ by the 3D VAE encoder, which performs spatiotemporal downsampling on a scale of $32 \times 32 \times 8$. Following this, the corresponding mask sequence is downsampled to the same scale, ensuring consistency. However, given that the 3D VAE encoder of \emph{LTX-Video} employs 3D Causal Convolutions, the temporal features consist of $1 + \frac{n-1}{8}$ frames. Consequently, in the temporal domain, the masks $M_{j}^{input} $ of eight temporally contiguous mask frames $M_{j}$ are intersected to form a single mask frame, except for the first mask frame. Ultimately, the input to the denoising video transformer combines the latent features of the masked video sequence with the spatiotemporally downsampled mask sequence, as shown in Figure \ref{fig:train_pipeline}. 

Furthermore, since the video containing the object to be removed is confined to a single scene, a brief prompt specific to that scene suffices. During the training stage, the pre-obtained prompt is encoded by the text encoder \cite{raffel2020exploring} to generate the prompt embeddings, which serve as an additional input for the denoising video transformer. Feature fusion between the prompt embedding tokens and the masked video feature tokens is achieved through cross-attention.

\begin{figure*}
	\centering
        \includegraphics[width=0.95\textwidth]{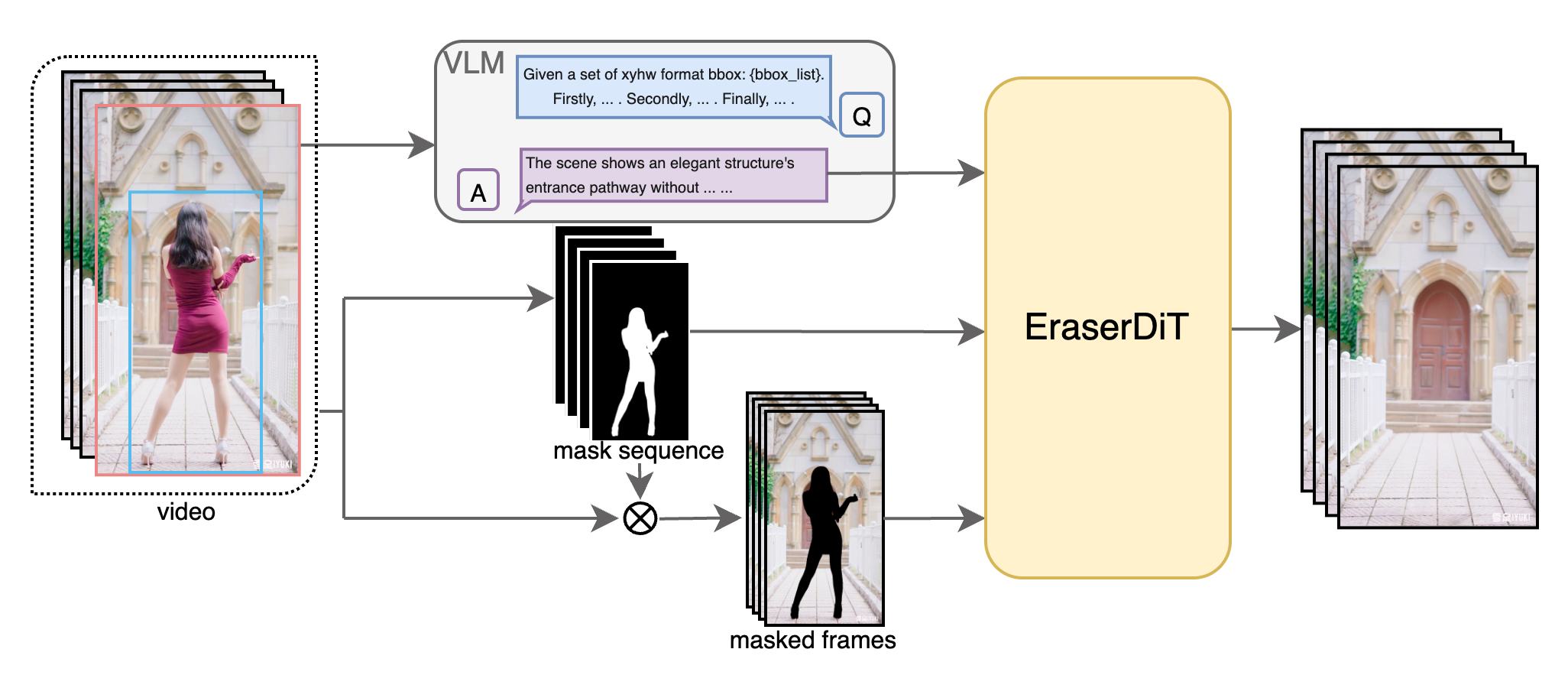}
	\caption{The inference pipeline of the proposed method for video object removal. As illustrated in the figure, in the process of automatically generating prompts that exclude the object to be removed using the VLM, ``Q'' and ``A'' represent fixed question formats and VLM's answer, respectively, used for interacting with the user and generating automated prompts that exclude the object to be removed.}
	\label{fig:test_pipeline}
\end{figure*}

\begin{figure}
	\centering
        \includegraphics[width=0.50\textwidth]{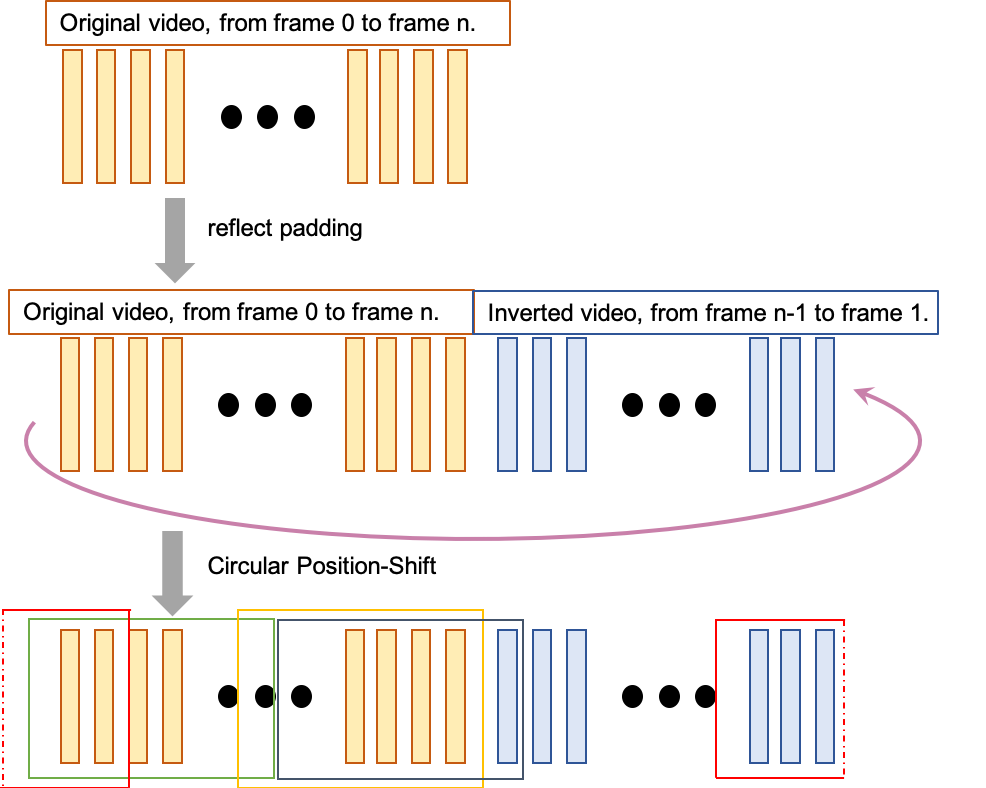}
	\caption{The Circular Position-Shift strategy involves reflect padding the original video sequence, which spans from frame 0 to frame n. After padding, the sequence extends from frame (n-1) to frame 1, and the two sequences are concatenated. Circular Position-Shift is then applied for inference.}
	\label{fig:test_circular}
\end{figure}

\begin{algorithm}
\caption{Circular Position-Shift for Long Sequence}
\label{alg:circular-position-shift}
\begin{algorithmic}[1]
\REQUIRE Masked video embedding $c_{mv}^{[0,l]}$ and mask map $c_{m}^{[0,l]}$ with length $l$, ,denosing steps $T$, initial nosiy latent $z_T^{[0,T]}$, pretrained EraserDiT model $ED(\cdot)$ for sequence length $f$, position-shift offset $\alpha$.
\ENSURE Denoised latent $z_0^{[0,l]}$.
\STATE Initialize accumulated shift offset $\alpha_{\sigma} = 0$
\STATE // Reflect sequence to construct circular sequence
\STATE $c_{mv}^{[0,l]+[l-1,1]}$, $c_{m}^{[0,l]+[l-1,1]}$, circular sequence length $L = 2l - 2$.
\STATE $indices = [0, l] + [l-1, 1]$

\FOR{$t = T$ \TO $1$}
    \STATE // Denoising loop
    \STATE Initialize start point $s=\alpha_{\sigma}$, end point $e=s+f$, processed length $pl=0$.
    \STATE // pad mode is warp
    \STATE $curr\_indices = right\_pad(indices, \alpha_{\sigma})$ 
    
    \WHILE{$pl < L$}
        \STATE $curr\_idx = curr\_indices[[s,e]]$
        
        \STATE $z_{t-1}^{curr\_idx} = ED(z_t^{curr\_idx}, c_{mv}^{curr\_idx}, c_{m}^{curr\_idx}, t)$
        
        $s \gets s + f$, $e \gets e + f$, $pl \gets pl + f$.
 
    \ENDWHILE
    \STATE $\alpha_{\sigma} \gets \alpha_{\sigma} + \alpha$
    
\ENDFOR
\RETURN Denoised latent $z_0^{[0,l]}$.
\end{algorithmic}
\end{algorithm}

\subsection{Inference pipeline}
During the testing stage, to enhance usability and achieve more precise video descriptions, we propose implementing an automated prompt method to generate text prompts. This approach ensures a more efficient and accurate alignment between the video content and the descriptive text. Therefore, we adopt the VLM method for video understanding and obtain its prompt. 

The complete inference pipeline is illustrated in Figure \ref{fig:test_pipeline}. First, the user selects a video sequence $V_{test}$ containing the object to be removed and marks the object with a bounding box in the first frame. Then, a vision-language model (VLM), such as Qwen2.5-VL~\cite{bai2025qwen2}, automatically generates a prompt describing the video as if the object were not present. In addition, based on the prompt provided by the user for the object to be removed, a mask sequence $M_{test}$ can be obtained. Therefore, the masked video sequence $V_{test}^{masked}$ is obtained by applying a pixel-wise multiplication of the video $V_{test}$ and the mask sequence $M_{test}$, that is $V_{test}^{masked} = V_{test} \ast  M_{test}$. At this stage, we have gathered all the necessary input conditions. By feeding these inputs into the model, we perform inference to obtain the final completed video $V_{out}$ without the target object.

To maintain consistency in the temporal domain, this paper proposes the Circular Position-Shift strategy for long sequences (as described in Figure~\ref{fig:test_circular} and Algorithm~\ref{alg:circular-position-shift}) that exceed the maximum training sequence length. In detail, we first pad the input sequence using the reflect mode. Next, we concatenate the beginning and the end to form a circular sequence. In a circular sequence, any selected subsequence is a physically coherent video segment. We set a cumulative shift offset, denoted as $\alpha_{\sigma}$, accumulating $\alpha$ at each timestep. This ensures that the model initiates the sliding window from different positions. Although using this strategy results in nearly double the inference cost, we employ the classifier-free guidance (CFG) distillation model to reduce the inference cost back to its original level. We distill the combined outputs of unconditional and conditional inputs into a single student model. Specifically, the student model is conditioned on a guidance scale and shares the same structures and hyperparameters as the teacher model. We initialize the student model with the same parameters as the teacher model and train it using a fixed guidance scale $3.0$. 

\section{Experiments}

\subsection{Training Data}
In the task of video object removal, obtaining paired training data (videos containing the target and corresponding ground truth without the target) is very challenging and unnecessary. Therefore, we will utilize synthetic data for training. This synthetic data encompasses a diverse range of background videos $V=\left \{ V_{1}, V_{2}, V_{3}, \dots ,V_{n}  \right \} $ and various masks $M=\left \{ M_{1}, M_{2}, M_{3}, \dots ,V_{n}  \right \} $. To achieve this, we downloaded 600,000 diverse background videos from the \emph{Pexels} website \footnote{https://www.pexels.com}. These videos encompass a variety of types, including landscapes, grasslands, buildings, mountains, and rivers. The mask data is composed of three components: masks from the open-source SA-V dataset \cite{ravi2024sam2segmentimages}, animal masks, and human masks. The latter two were obtained by crawling the internet for videos of animals moving and people dancing videos, whose masks are extracted by SAM 2 \cite{ravi2024sam2segmentimages}. 

\textbf{Animal masks.} We first downloaded video clips of large animals from the material video website. Since these videos generally lack scene changes, we extracted the masks according to their categories. Ultimately, we obtained approximately 30,000 videos, each with a duration ranging from 1 to 5 seconds.

\textbf{Human masks.} Firstly, we need to download the raw videos containing people from the Internet, due to the characters lacking diversity in the material video website. To achieve more accurate tracking of the specified person and obtain their mask in the video, we exclusively retain single-person video clips through Object Detection. To ensure the continuity of target movement, it is crucial to avoid scene transitions within video clips. Initially, we utilized PySceneDetect for preliminary scene segmentation due to its speed, despite its lower accuracy. After that, for a more precise extraction of video clips without scene transitions or target switches, we employ manual annotation to obtain the final video clips. Finally, we tracked the human and extracted its masks from all video clips using Grounded SAM 2 \cite{ren2024groundedsam} and saved them. We processed approximately 32,000 videos, each with a duration ranging from 5 to 20 seconds.

\subsection{Training Details}
% We utilize \emph{LTX-Video} \cite{hacohen2024ltx} as the foundational architecture. We only train the parameters of the video transformer model for video inpainting, whose training details are described in Section \ref{sec:video-transformer}. However, we found the presence of severe artifacts after the 3D VAE of \emph{LTX-Video} \cite{hacohen2024ltx}, so we finetune the parameters of the VAE Decoder and freeze the Encoder's, whose fine-tune details are described in Section \ref{sec:vae-decoder}.

% \begin{figure*}
% 	\centering
%         \includegraphics[width=0.99\textwidth]{figs/vae_demo_1.png}
% 	\caption{The results of VAE compared with \emph{LTX-Video}. In this figure, from the first column to the last column, the images display the input, the result of \emph{LTX-Videp} VAE, and our finetuned VAE, respectively.}
% 	\label{fig:vae_demo}
% \end{figure*}

% \subsubsection{Video Transformer}
% \label{sec:video-transformer}
We utilize \emph{LTX-Video} \cite{hacohen2024ltx} as the foundational architecture and only train the parameters of the video transformer model for video inpainting.
The video transformer is trained with a resolution of $960 \times 960$ and 81 frames. The training process comprises 280k iterations on 24 NVIDIA A100 (80G) GPUs, utilizing the \emph{Adam-W} optimizer with an initial learning rate of $3 \times 10^{-5}$. In addition, to achieve faster and more stable training, the \emph{Rectified Flow} \cite{lipman2022flow} is employed. The loss function is implemented in two distinct stages. In the first stage, spanning the initial 150k iterations, we exclusively utilize the L2 loss function to optimize the video transformer. To enhance the handling of video object removal with significant motion, the frame sampling step is randomly chosen from 1 to 6 during the training phase.

Subsequently, in the second stage, we shift our focus to the masked region by employing the Focal Area loss function for the remaining iterations. Focal Area loss is defined by
\begin{equation}
    L_{focal} = L_{2} * (1 + D_{mask}),
\end{equation}
and the $D_{mask}$ is the dilated mask pf $M_{j}^{input}$.

\subsection{Ablation Study on Temporal Consistency}
To evaluate the effectiveness of the proposed Circular Position-Shift (CPS) mechanism in maintaining temporal consistency across long video sequences, we conducted an ablation study comparing variants with and without CPS. As temporal coherence is best observed in dynamic contexts, we provide video comparisons in the supplementary material to highlight the differences. Without CPS, the model exhibits noticeable frame-wise inconsistencies and texture flickering, especially in repetitive motion patterns and gradual transitions. In contrast, incorporating CPS significantly improves the stability of object boundaries and background continuity throughout the sequence, demonstrating its advantage in preserving temporal coherence over extended durations.
% \subsubsection{VAE Decoder}
% \label{sec:vae-decoder}
% As illustrated in Figure \ref{fig:vae_demo}, the original VAE of LTX-Video exhibits chessboard artifacts in scenarios involving significant motion, particularly evident in the final frame. Therefore, we fine-tuned the decoder section of the VAE. In addition to the commonly used $L_1$ reconstruction loss, we integrate perceptual loss $L_{dists}$ and GAN adversarial loss $L_{adv}$ to further improve the quality of reconstruction. The complete loss function is shown in Equation~\ref{eq:vae_loss}.

% \begin{equation}\label{eq:vae_loss}
%     Loss_{vae}=L_{1}+L_{dists}+0.05L_{adv}
% \end{equation}

\subsection{Performance Evaluation}

\begin{figure*}
	\centering
        \includegraphics[width=0.99\textwidth]{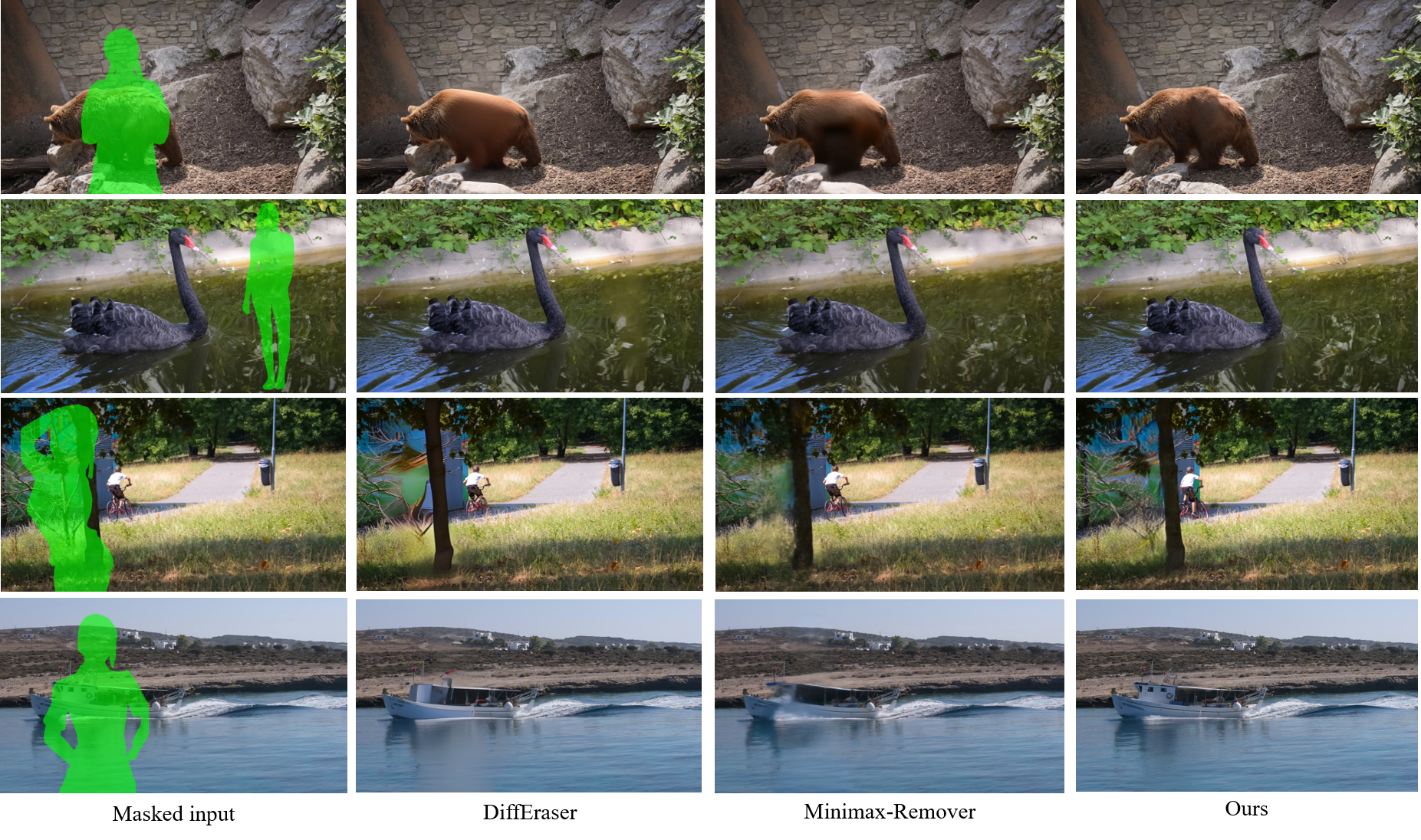}
	\caption{The results on the DAVIS dataset \cite{perazzi2016benchmark} are presented as follows: the first column shows the input masked frames, the second to fourth columns display the results of DiffEraser~\cite{li2025diffueraser}, Minimax-remover~\cite{zi2025minimax}, and ours.}
	\label{fig:result_1}
\end{figure*}

\begin{figure*}
	\centering
        \includegraphics[width=0.99\textwidth]{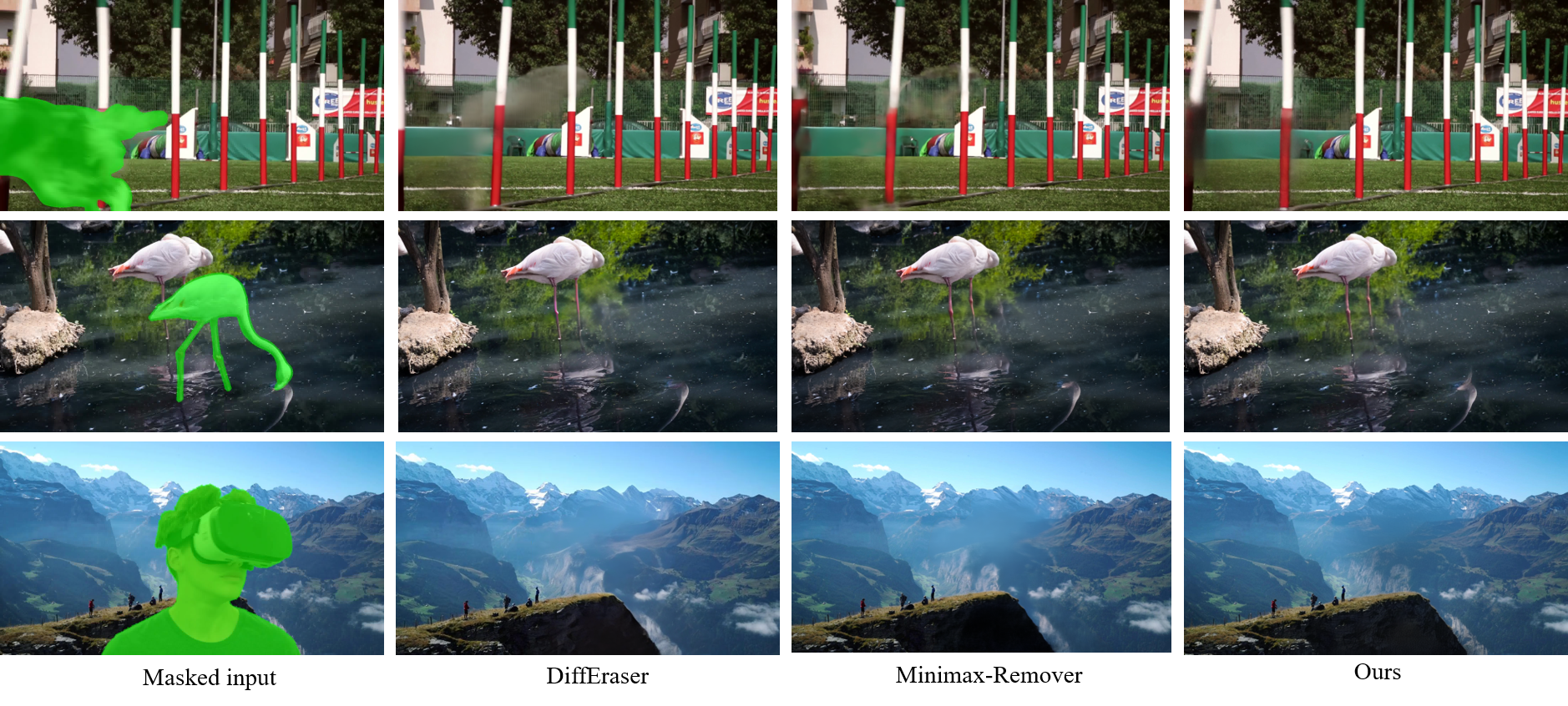}
	\caption{The results on the DAVIS \cite{perazzi2016benchmark} and HQVI~\cite{cho2025elevating} dataset are presented as follows: the first column shows the input masked frames, the second to fourth columns display the results of DiffEraser~\cite{li2025diffueraser}, Minimax-remover~\cite{zi2025minimax}, and ours.}
	\label{fig:result_2}
\end{figure*}

% \begin{figure*}
% 	\centering
%         \includegraphics[width=0.99\textwidth]{figs/davis_results.png}
% 	\caption{The results on the DAVIS dataset are presented as follows: the first column shows the input masked frames, the second column displays the results of ProPainter, and the third column contains our results.}
% 	\label{fig:result_davis}
% \end{figure*}

% \begin{figure}
% 	\centering
%         \includegraphics[width=0.50\textwidth]{figs/results_1_1.png}
% 	\caption{The results of ProPainter and the proposed method. There are more textures and fewer artifacts in the proposed method.}
% 	\label{fig:result_1}
% \end{figure}

% \begin{figure}
%         \centering
%         \includegraphics[width=0.50\textwidth]{figs/results_2_t_1.png}
% 	\caption{The results of ProPainter and the proposed method. They demonstrate the context consistency in the temporal domain.}
% 	\label{fig:result_2}
% \end{figure}

% Table
\begin{table*}%
% \caption{Quantitative results on the DAVIS \cite{perazzi2016benchmark} and HQVI~\cite{cho2025elevating} dataset. "Time" denotes the average inference time per video with a resolution of 2160 × 1200 with 121 frames (\emph{DiffEraser} is tested on the resolution of $960 \times 528$) without any acceleration method on H800.}

\begin{minipage}{\columnwidth}
\begin{center}
\begin{tabular}{l|lllll|llll}
  % TERRAIN\footnote{This is a table footnote. This is a
  %   table footnote. This is a table footnote.}   & (200m$\times$200m) Square\\ \midrule
  \hline
  \hline
  \multirow{2}{*}{Method} & \multicolumn{5}{c|}{HQVI} & \multicolumn{4}{c}{DAVIS} \\

  & PSNR$\uparrow$ & SSIM $\uparrow$ & LPIPS $\downarrow$ & VFID $\downarrow$ & Time $\downarrow$ & PSNR$\uparrow$ & SSIM $\uparrow$ & LPIPS $\downarrow$ & VFID $\downarrow$ \\ 
  \hline
  DiffEraser       & 26.89 & 0.8480 & 0.0623 & 0.0502 & \textcolor{red}{73s (960*528) }& 27.61& 0.8234&0.0891 &0.0573 \\
  MiniMaxRemover   & 27.41 & 0.8772 & 0.0877 & 0.0174 & 7min20s &27.63 &0.8819 &0.1139 &0.0219\\
  % RGVI             & 30.10 & 0.9489 & 0.0357 & \textbf{0.0058} & -& & & &\\
  Ours(EraserDiT)  & \textbf{30.78} &\textbf{0.9446} & \textbf{0.0343} & \textbf{0.0072} & \textbf{65s} &\textbf{31.10}&\textbf{0.9488} &\textbf{0.0409} &\textbf{0.0114}\\
  \hline
  \hline
\end{tabular}
\end{center}
\bigskip\centering
% \footnotesize\emph{Source:} This is a table
%  sourcenote. This is a table sourcenote. This is a table
%  sourcenote.

 % \emph{Note:} This is a table footnote.
\end{minipage}
\caption{Quantitative results on the DAVIS \cite{perazzi2016benchmark} and HQVI~\cite{cho2025elevating} dataset. "Time" denotes the average inference time per video with a resolution of 2160 × 1200 with 97 frames (\emph{DiffEraser} is tested on the resolution of $960 \times 528$) without any acceleration method on H800.}
\label{tab:metric}
\end{table*}%

\textbf{Visual results.} 
The qualitative comparisons are illustrated in Figure \ref{fig:result_1} and Figure \ref{fig:result_2}. While \emph{DiffEraser} and \emph{Minimax-Remover} often produce completion results with pronounced artifacts and structural distortions, our method yields visually coherent reconstructions with minimal artifacts. Notably, it adeptly restores complex natural textures, such as water surfaces and tree foliage, resulting in outputs that exhibit enhanced detail and perceptual realism. Upon closer inspection, our method excels at synthesizing textures that harmonize with the surrounding background, especially during the removal of large foreground objects. In contrast, both \emph{DiffEraser} and \emph{Minimax-Remover} tend to introduce unnatural patterns into the completed regions, disrupting visual consistency.

Moreover, in areas featuring intricate patterns—such as striped surfaces or dynamic water ripples—\emph{DiffEraser} and \emph{Minimax-Remover} struggle to replicate the inherent visual complexity and continuity. Our approach, by contrast, demonstrates superior capability in modeling these regions, effectively capturing the fluid dynamics and structural repetition present in the original footage.

Beyond the issues previously discussed, our experiments reveal additional limitations in baseline methods when evaluating long-term temporal coherence. Specifically, the Differaser baseline exhibits pronounced jittering and flickering artifacts across frames, undermining the visual stability of generated sequences. This effect is especially noticeable during smooth camera motions and subtle temporal transitions. On the other hand, MinimaxRemover demonstrates aggressive foreground suppression, often leading to unnatural removal of object boundaries and erosion of fine-grained details. These issues are best appreciated in motion, and we provide corresponding video comparisons in the Project Page to illustrate the temporal inconsistencies and structural degradation observed in these baselines.

% \textbf{Temporal results.} In the temporal domain, our results maintain good consistency throughout the entire video sequence compared to \emph{DiffEraser} and \emph{Minimax-Remover}, as shown in Figure \ref{fig:result_2}. The bridge completed by ProPainter is not straight and exhibits ghosting, as indicated by the red box in the second row of Figure \ref{fig:result_2}. In contrast, our method produces a consistent bridge in each frame, as seen in the last row of Figure \ref{fig:result_2}.  

% \textbf{Evaluation on DAVIS.} 
% The visual results on DAVIS \cite{perazzi2016benchmark} and HQVI~\cite{cho2025elevating} datasets are shown in Figure \ref{fig:result_1} and \ref{fig:result_2}. Upon observation, it becomes evident that our method excels in reconstructing textures consistent with the background during the removal of large objects, while the completed areas of \emph{DiffEraser} and \emph{Minimax-Remover} are marred by numerous unnatural artifacts. Moreover, in regions characterized by patterns such as stripes or water ripples,  \emph{DiffEraser} and \emph{Minimax-Remover} struggle to accurately capture their natural states. In contrast, our method effectively represents these natural patterns and the dynamics of water movement.

\textbf{ Quantitative Results.}
Besides, we generate binary masks and reconstruct the masked videos using the DAVIS dataset \cite{perazzi2016benchmark}, while directly leveraging the ground-truth masks provided in the HQVI dataset \cite{cho2025elevating}. This setup enables the computation of quantitative performance metrics. As widely acknowledged, video completion tasks place greater emphasis on visual fidelity. Therefore, we employ PSNR, LPIPS, SSIM, and VFID to comprehensively evaluate the perceptual quality and temporal consistency of the completed videos. As presented in Table \ref{tab:metric}, our proposed method consistently outperforms both \emph{DiffEraser} and \emph{Minimax-Remover} across all considered evaluation metrics.

Furthermore, during inference, as detailed in Table \ref{tab:metric}, the proposed approach completes a video of resolution $2160 \times 1200$ with 97 frames using 40 denoising steps in just 65 seconds on a single NVIDIA H800 GPU. In contrast, \emph{Minimax-Remover} requires 7 minutes and 20 seconds to process the same input with only 6 denoising steps on the same hardware. Due to its resolution limitation of $960 \times 528$, \emph{DiffEraser} was evaluated at its maximum supported resolution, completing the same sequence in 73 seconds. These results demonstrate that our method achieves significant speedup while preserving superior visual quality, highlighting its efficiency and effectiveness in high-resolution video completion tasks.

% % Table
% \begin{table}%
% \caption{Quantitative comparisons on DAVIS datasets}
% \label{tab:one}
% \begin{minipage}{\columnwidth}
% \begin{center}
% \begin{tabular}{llll}
%   % TERRAIN\footnote{This is a table footnote. This is a
%   %   table footnote. This is a table footnote.}   & (200m$\times$200m) Square\\ \midrule
%   Method & SSIM $\uparrow$ & LPIPS $\downarrow$ & FVD $\downarrow$ \\ 
%   ProPainter     & 0.9492 & 0.0379 & 120.53\\
%   Ours(EraserDiT)  & \textbf{0.9673} &\textbf{0.0320} & \textbf{87.08}\\
% \end{tabular}
% \end{center}
% \bigskip\centering
% % \footnotesize\emph{Source:} This is a table
% %  sourcenote. This is a table sourcenote. This is a table
% %  sourcenote.

%  % \emph{Note:} This is a table footnote.
% \end{minipage}
% \end{table}%

\section{Limitations}
Due to the inherent limitations of the current video generation fundamental mode, the results are suboptimal in scenarios involving rapidly flowing water and extremely fast camera movements. In future work, our aim is to address these challenges by incorporating alternative fundamental modes of video generation to improve performance in such complex scenes.

\section{Conclusion}
In this work, we present a novel video object removal framework based on the DiT (Denoising Diffusion Transformer) model, enabling precise and effective erasure of large objects in high-resolution video sequences. To support scalable and accurate object localization, we design a specialized data preprocessing pipeline for generating large object masks, and propose an automated prompt generation strategy for handling videos that do not contain the target object. These components enhance robustness across diverse scenarios and improve adaptability during inference. Furthermore, we introduce a Circular Position-Shift (CPS) mechanism to better maintain long-range temporal consistency during video generation. The CPS design effectively aligns frame-wise representations across extended sequences, reducing temporal flickering and preserving structural continuity. Quantitative and qualitative results show that our method achieves state-of-the-art performance in texture fidelity, detail recovery, and consistency, setting a new benchmark for high-resolution video inpainting tasks.

\bibliography{aaai2026}

\end{document}